\documentclass[11pt]{article}

\usepackage[preprint]{acl}
\usepackage{times}
\usepackage{latexsym}

\usepackage[T1]{fontenc}
\usepackage[utf8]{inputenc}

\usepackage{microtype}

\usepackage{inconsolata}

\usepackage{graphicx}
\usepackage{mathtools}
\usepackage{amsmath}
\usepackage{listings}
\title{Can LLMs Narrate Tabular Data? An Evaluation Framework for Natural Language Representations of Text-to-SQL System Outputs}
\author{
\textbf{Jyotika Singh\textsuperscript{}},
\textbf{Weiyi Sun\textsuperscript{}},
\textbf{Amit Agarwal\textsuperscript{}},
\textbf{Viji Krishnamurthy\textsuperscript{}},\\
\textbf{Yassine Benajiba\textsuperscript{}},
\textbf{Sujith Ravi\textsuperscript{}},
\textbf{Dan Roth\textsuperscript{}}
\\ 
\\
 \textsuperscript{}Oracle AI
\\
 \small{
   \textbf{Correspondence:} \href{mailto:email@domain}{jyotika.s.singh@oracle.com}
 }
}

\begin{document}
\maketitle
\begin{abstract}
In modern industry systems like multi-turn chat agents, Text-to-SQL technology bridges natural language (NL) questions and database (DB) querying. The conversion of tabular DB results into NL representations (NLRs) enables the chat-based interaction. Currently, NLR generation is typically handled by large language models (LLMs), but information loss or errors in presenting tabular results in NL remains largely unexplored.
This paper introduces a novel evaluation method - Combo-Eval - for judgment of LLM-generated NLRs that combines the benefits of multiple existing methods, optimizing evaluation fidelity and achieving a significant reduction in LLM calls by 25-61\%. Accompanying our method is NLR-BIRD, the first dedicated dataset for NLR benchmarking. Through human evaluations, we demonstrate the superior alignment of Combo-Eval with human judgments, applicable across scenarios with and without ground truth references. 
\end{abstract}

\section{Introduction}

In real-world Natural Language Processing (NLP) applications built around Large Language Models (LLMs) \citep{Khurana2022, vaswani2023attentionneed}, there has been a growing prominence of communicating using plain text across diverse data types \citep{soudani2024surveyrecentadvancesconversational, duan-etal-2024-botchat, liu-etal-2022-augmenting-multi}. Natural language (NL) interfaces to databases (DBs) are increasingly becoming integral in industry applications~\citep{Singh2023} and agent workflows, leveraging Text-to-SQL systems to convert user questions into structured SQL queries and subsequently presenting query execution results (tables) in an NL format \citep{azure_video, google_analytics}. This transition from raw tables to NL responses is critical, transforming the impersonal “Computer says no” into user-friendly interactions that enhance accessibility and engagement \citep{blog_quote}.

To build conversational applications for DBs, two core components are essential: the generation of SQL queries based on user questions, and the vernacular presentation of the DB outputs, the latter being the focus of our study. Traditionally, the emphasis has been on enhancing SQL generation accuracy \citep{iacob-etal-2020-neural}, leaving a gap in methodologies for evaluating natural language representations of SQL execution results (NLRs). Our research addresses this gap by evaluating LLMs in the task of transforming SQL execution result-sets into natural language but also by proposing enhanced methodology for such evaluations.

Existing evaluations, such as those discussed by \citet{wang-etal-2024-revisiting}, highlight limitations of metrics alone in assessing complex NL tasks. While metrics are not entirely sufficient, our findings suggest they offer valuable signals when paired with LLM-based evaluations. We propose a composite method, Combo-Eval, which combines metrics with LLM-as-a-judge to enhance agreement with human evaluations while significantly reducing computational overhead.

Our contributions include:
\begin{itemize}
    \item \textbf{Combo-Eval Method:}
    An evaluation technique that synergizes metrics with LLM-assessment, reducing LLM calls while maintaining high evaluation fidelity.
    \item \textbf{NLR-BIRD Dataset:}
    A new NLR-BIRD dataset spanning 11 domains, enabling evaluation of NLR generation in context of Text-to-SQL systems.
    \item \textbf{Comprehensive Benchmarking:} Evaluation of
    %\item Comprehensive benchmarking of 
    15 judge LLMs and multiple scenarios on the task of NLR judging. We also share results across multiple LLMs on the task of NLR generation and share challenges.
    \item \textbf{Evaluation Scenarios:} 
    Comparative analysis with and without ground-truth references to emulate real-world industry applications where references may be unavailable.
\end{itemize}

\section{Background and Related Work}
Research in natural language interfaces to DBs has predominantly focused on two areas: Text-to-SQL conversion \citep{iacob-etal-2020-neural} and long-form table question answering (LFTQA) \citep{Eratta, zhao-etal-2024-tapera, nan-etal-2022-fetaqa}. 

Text-to-SQL targets SQL generation complexities while neglecting the NLR phase critical for user-interactive systems. Emerging studies have begun to explore NL renditions from DB outputs, yet they only scratch the surface with no mentions of evaluation of these NLRs \citep{https://doi.org/10.48550/arxiv.2408.14717}. 

Unlike LFTQA, which emphasizes gleaning answers directly from tables, where answers may be a part of the available table, our work centers on narrating complete tabular datasets following SQL query execution. This narrative task demands accurately conveying full table data in natural language, a requirement particularly relevant for Text-to-SQL implementations in conversational systems.

Methods for evaluating data similar to NLRs have been explored in LFTQA \citep{wang-etal-2024-revisiting}, indicating metrics as a poor judge that fail the evaluation task, and showing a preference for LLM-based evaluation. 
In our work, we developed the Combo-Eval method that outperforms singular LLM-judges by enhancing utility of metrics and integrating metric-based thresholds.

Despite assumptions that simpler tasks like table-to-NL conversion would be straightforward, our findings reveal that LLMs face significant hurdles even in this domain. Our primary focus in this paper is on evaluation of NLRs generated from DB results across evaluation techniques and models.

More details on related work is in Appendix~\ref{sec:relatedwork}

\section{NLR Dataset}
\label{sec:nlrdatasetmain}
We introduce NLR-BIRD\footnote{\url{https://sites.google.com/view/nlr-bird/home}} dataset which contains NLRs across questions present in the BIRD-dev dataset \citep{li2023llmservedatabaseinterface}, presenting a new data point for NLR, thereby enabling users to test components of DB interaction systems end-to-end, across 11 domains – financial, debit card specializing, formula 1, codebase community, European football 2, student club, California schools, card games, superhero, toxicology, and thrombosis prediction. 
The data collection process is shared in Appendix~\ref{sec:datacuration}.

The BIRD dataset contains 1534 NL questions paired with ground truth SQL queries and a BIRD DB. We used row count (rc) and column count (cc) to measure the result size after executing these queries on the DB.
Large result sets may benefit from a summarization strategy to formulate the NLR for meaningful rendering and readability. The summary method may depend on the application: some needing a general summary, others requiring partial data representation.
To keep a consistent style, large result sets (rc+cc$\ge$500), comprising 4.3\% of samples (Figure~\ref{fig:dataset_result_analysis} in Appendix~\ref{sec:datattrs} shows the distribution of result size in BIRD dataset), were excluded. The NLR-BIRD dataset includes 1468 samples across 11 domains, containing human-labeled ground truth NLR for rc+cc<500 that narrates the tabular response generated by SQL execution.

\begin{figure}[t]
	\includegraphics[width=\columnwidth]{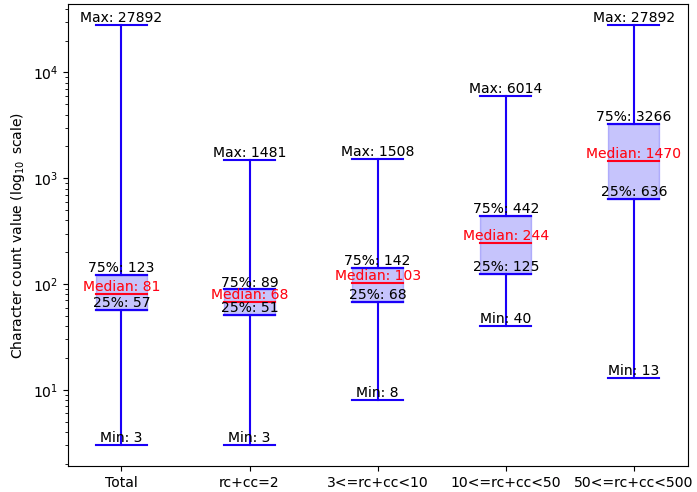}
	\caption{Box plot of percentiles for NLRs in the dataset representing character counts across result-set sizes.}
	\label{fig:boxplot-chars}
\end{figure}

Figure~\ref{fig:boxplot-chars} depicts the correlation between the length of NLRs and the result size, indicating a clear upward trend in character count as rc+cc increases. More comprehensive statistics, including word counts, are available in Appendix~\ref{sec:datattrs}.

The maximum length of any word in the NLRs is 92 characters, which corresponds to a URL. The NLRs contain 19601 characters corresponding to numbers and 62140 characters corresponding to alphabets.

\section{Methods}
This section describes the three evaluation methods and the two scenarios in which the methods are applied. One scenario assumes the presence of ground truth (GT) and uses it as a reference to assess the evaluation strategies. The other scenario handles cases where GT is unavailable, employing source information as a proxy for ground truth.

Given the straightforward nature of transforming simple result-sets (rc=1 and cc=1; total rc+cc=2) into NLRs due to limited information context, our experiments and evaluations emphasize result-sets with rc+cc$\geq$3. We sampled our dataset to represent the different result size categories uniformly.
Human evaluators assessed the correctness (labeled 0 or 1) of NLRs (using the same process described for dataset curation in appendix~\ref{sec:datacuration}) generated by four LLMs for each sample: Phind-CodeLLma-34B-v2, Llama-3.1-70B-Instruct, Llama-3.1-405B-Instruct \citep{touvron2023llamaopenefficientfoundation}, and GPT-4o \citep{gpt4o}. This evaluation yielded 660 labeled NLRs corresponding to 165 unique questions, identifying class 0 as incorrect and class 1 as correct.

The above LLM-generated human-assessed NLRs serve as benchmarks to judge different evaluation methods (described in Section~\ref{sec:evalmeths}). We assess per-class recall, precision, F1 score, as well as macro recall, precision, F1, and overall accuracy. In scenarios with significant class imbalances, macro-averaging is favored to aptly capture the minority class 0, crucial for industry applications where inaccuracies can detrimentally impact system utility. We present the F1 scores in the main paper and include other scores in the Appendix. 

While human-assessed NLRs served as benchmarks to evaluate different methods, we designated 25\% of the samples (generated by GPT-4o) as the dev set, which informed our threshold calibration, prompt engineering, and inference parameters. The remaining samples, generated by other models, formed the test set. Results presented hereafter pertain to evaluations on the test set. Details about the experimental setup are in Appendix~\ref{sec:prompts} and~\ref{sec:params}.

\subsection{Evaluation Scenarios}

We present two scenarios in which our evaluation methods are applied.

1) Model Generated NLR compared to Ground-Truth NLR (\textbf{GT}): This approach uses the NLRs from the NLR-BIRD dataset as the reference, assessing the model-generated NLRs against annotated ground-truth NLRs.

2) Model Generated NLR compared to User Question and DB Result-Set (\textbf{UQDB}): This is derived by appending the user’s question and DB result-set table for the reference text. This does not rely on annotated ground truth NLRs and relies on source information instead.

\subsection{Evaluation Methods}
\label{sec:evalmeths}

1) \textbf{Metrics-as-a-Judge}: Metrics including cosine similarity, BERTScore \citep{2020BERTScore}, and ROUGE scores \citep{lin-2004-rouge} were considered. A threshold was determined on the dev set to make the decision boundary. Further details can be found in Appendix~\ref{sec:evalsmetrics} and ~\ref{sec:Thresholds}.

2) \textbf{LLM-as-a-judge}: Automated evaluation using LLM-judge \citep{https://doi.org/10.48550/arxiv.2411.15594,raina-etal-2024-llm}

3) \textbf{Combo-Eval} method combines the above two approaches in an attempt to retain benefits of both to attain superior evaluation along with efficiency advantages by limiting the number of LLM calls. 

\begin{figure}[t]
	\includegraphics[width=\columnwidth]{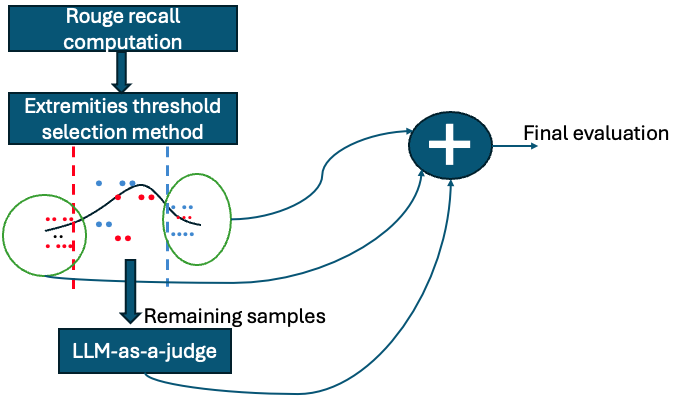}
	\caption{The Combo-Eval method flow combining metrics-based evaluation and LLM-as-a-judge.}
	\label{fig:Combo-Eval}
\end{figure}

We started with computing metrics and determined upper and lower thresholds for each class (more details in Appendix~\ref{sec:Thresholds}). The samples that didn't pass through the extremity threshold were sent to LLM-judge for a finer diagnostic. Figure~\ref{fig:Combo-Eval} shows the flow of this method. To represent this mathematically, let’s denote:

\(R\) (ROUGE-1 recall); \(C\) (class label 0 or 1)

\(L\) (Output of LLM-as-a-judge – True for class 1 and False for class 0)

\(th_0l\) - \(th_0u\) are lower and upper thresholds for class 0, and \(th_1l\) - \(th_1u\) are for class 1.

The classification based on ROUGE score can be represented as:   

\(C = 1\) if \(th_11 < R =< th_1u\)

\(C = 0\) if \(th_0l < R =< th_0u\)

\(C = pending\) otherwise

Then, the final evaluation can be represented using mathematical notation as 
\[C = 
\begin{cases} 
	1 & \text{if } th_1l < R \leq th_1u \\
    & \quad \vee \left( (th_0u < R \leq th_1l \right) \\
    & \quad \vee \left(R \leq th_0l \right) \vee \left (R > th_1u)\right) \land L \\
	& \\
	0 & \text{if } th_0l < R \leq th_0u \\
	& \quad \vee \left( (th_0u < R \leq th_1l \right) \\
	& \quad \vee \left(R \leq th_0l \right) \vee \left (R > th_1u)\right) \land \neg L \\
\end{cases}\]

where \(\vee\) represents logical OR, \(\land\) represents logical AND, and \(\neg\) represents logical NOT operation.

\section{Results and Discussions}
\label{sec:resultsanddis}
\subsection{NLR Generation Model Performance}

Quality of LLM-generated NLRs decreases significantly as the result size increases. Table~\ref{tab:llmeval} details human evaluation of LLM-generated NLRs, segmented by result size. It reveals a consistent trend where LLMs perform better with smaller result-sets, with performance decreasing 10-30\% as result-set size increases from <10 to $\ge$10. Specifically, larger models like Llama-3.1-405B-instruct generally outperform Phind-CodeLlama-34B-v2, Llama-3.1-70B-instruct, and GPT-4o, particularly on challenging result-set sizes >=10. Notably, GPT-4o excels in the smallest size category of <10 but underperforms with larger sizes.

\begin{table}
    \begin{center}
        \begin{tabular}{|p{1.6cm}|p{0.9cm}|p{0.9cm}|p{0.9cm}|p{0.9cm}|}
            \hline
            Result size & Phind & L3.1 70 & L3.1 405 & GPT-4o \\
            \hline
            3-9 & 0.80 & 0.87 & 0.89 & 0.91\\
            10-49 & 0.60 & 0.77 & 0.83 & 0.62\\
            50-499 & 0.52 & 0.57 & 0.59 & 0.30\\
            \hline
            Total & 0.65 & 0.75 & 0.78 & 0.63\\
            \hline
        \end{tabular}
        \caption{Human evaluation of NLRs generated by LLMs - percentage of questions where NLR generated by LLM is rated as acceptable by human evaluators. Phind=Phind-CodeLlama-34B-v2; L3.170=Llama 3.1-70B-instruct; L3.1405=Llama 3.1-405B-instruct.}
        %Result size = rc+cc (defined in Section~\ref{sec:nlrdatasetmain})}
        \label{tab:llmeval}
    \end{center}
\end{table}

\begin{figure}[t]
	\includegraphics[width=\columnwidth]{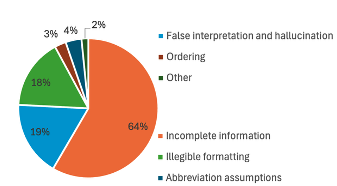}
	\caption{Reasons for incorrect NLRs based on human assessment of model-produced NLRs.}
    \label{fig:inaccuracy_reasons_overall}
\end{figure}

The most common cause of incorrect NLRs was incomplete information. As indicated in Figure~\ref{fig:inaccuracy_reasons_overall}, common reasons for incorrect NLRs included missing elements from the result-set rendering the outcome incomplete, hallucinations, rendering results out of order for questions where the order was important, skipping nulls rather than indicating the value was unavailable, formatting inconsistencies impacting readability, and more. Incorrect NLRs at times had inconsistent behavior in de-duplicating result-set values, and inconsistent behavior in interpreting ambiguities in user question or result-set. 
All identified inaccuracies stem from the LLM generation of NLR, further explained in Appendix~\ref{sec:err}. A breakdown of issues across different LLMs and examples of correct and incorrect NLRs are shared in Appendix~\ref{sec:err}.

Formatting inconsistencies, particularly in GPT-4o outputs, were prominent with increased result sizes. Discrepancies at the beginning versus the end of NLR text and spacing issues were pronounced, as illustrated in a detailed example in Appendix~\ref{sec:gptformateg}.

While smaller, structured result-sets are more amenable to LLM processing, larger complex tables pose more challenges. These may exceed LLM context windows, resulting in verbose or impractical NLR outputs for users.

\subsection{Comparison Between Evaluation Methods}
We evaluated model generated NLRs using three methods (metric-based, LLM-as-a-judge, and Combo-Eval) to analyze how well these methods align with human judgments of NLR completeness/comphrehensiveness, faithfulness, and readability. Our results indicate that Combo-Eval exhibits highest alignment with human judgment.

\begin{figure}[t]
	\includegraphics[width=\columnwidth]{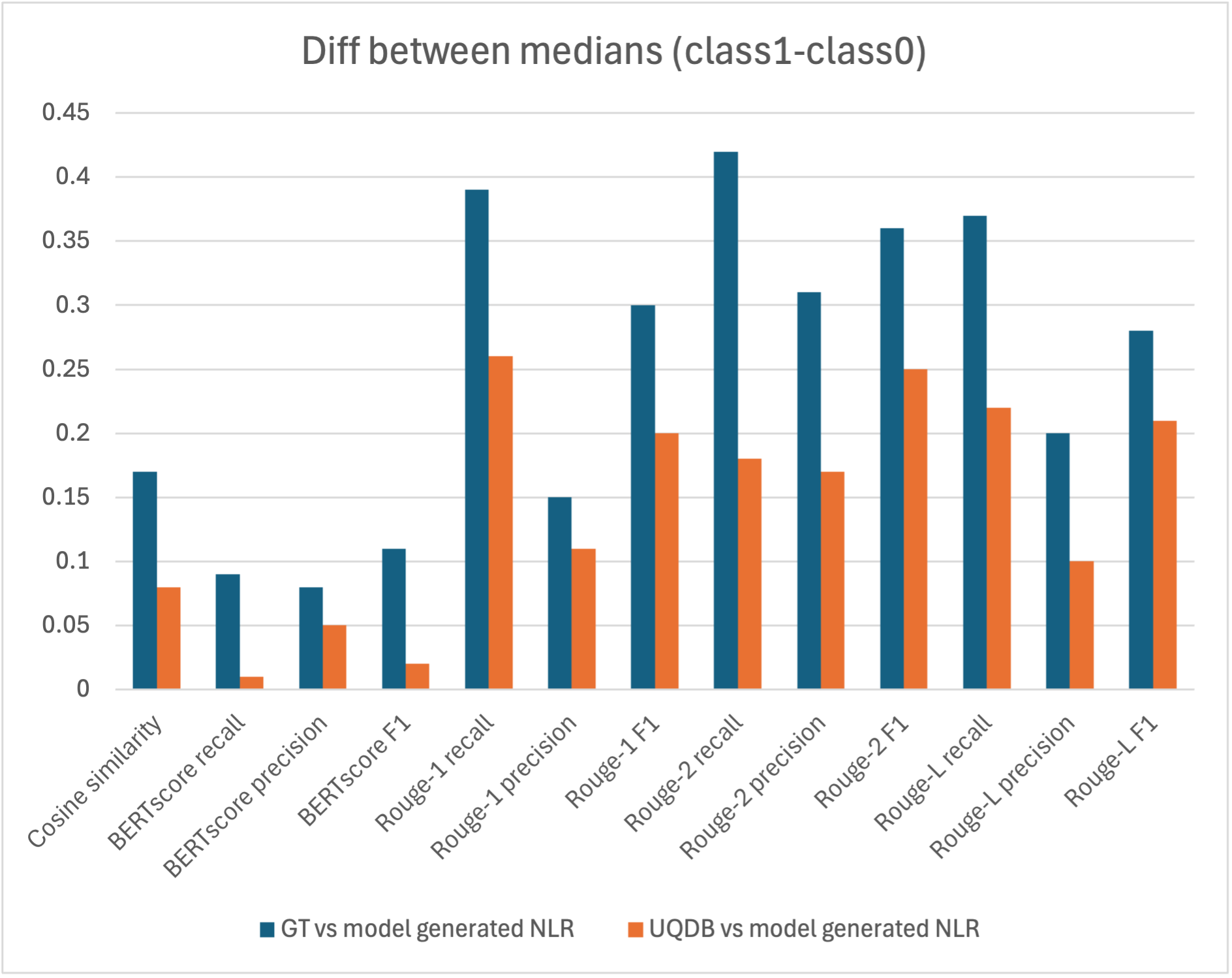}
	\caption{Difference between median scores of class 1 and class 0. Scores are computed between model generated NLRs and (GT (blue) \& UQDB (orange).)}
	\label{fig:median_diffs_2}
\end{figure}

\paragraph{Metrics-as-a-Judge:}
Recall-based measures, such as ROUGE, measure stronger differences between correct and incorrect NLRs, compared to other metrics. We calculated various automated metrics between model-generated NLRs against GT and UQDB. As shown in Figure~\ref{fig:median_diffs_2}, there is a notable differentiation in median metric scores for class 1 (correct NLRs) and class 0 (incorrect NLRs). For a problem of this nature, we anticipate observing a higher frequency of word/n-gram matches for effective NLRs. This is because tabular data retrieved from a query often reflects specific values within tables, which may be more appropriately rendered as-is in the NLR, rather than substituting them with semantically similar alternative words. This is likely why BERTscore seems to be comparatively less prominent.

The model-generated NLRs with GT exhibit the most difference in medians between correct and incorrect NLRs, indicating a trend that could aid in evaluating correct versus incorrect NLRs. UQDB follows a similar pattern but with a relatively smaller difference.

We use ROUGE-1-recall metric with a threshold as decision boundary to classify NLRs into correct vs. incorrect based on general applicability of this metric and the maximum median difference between class 1 and 0. Decision boundaries and results for all the metrics are shared in Appendix~\ref{sec:evalsmetrics}.

Table~\ref{tab:eval} reveals that metrics can help in evaluating NLR correctness, although identifying incorrect NLRs remains challenging based on per-class scores. 
In real-world systems, correctly identifying
incorrect NLRs is important and opens opportunities to put correction/flagging measures in place. More statistics and scores are in Appendix~\ref{sec:gpt4oresdets}.

\begin{table*}
	\begin{center}
	\begin{tabular}{|c|c|c|c|c|c|c|c|c|c|c|c|}
    \hline
         \multicolumn{6}{|c|}{GT} &  \multicolumn{6}{c|}{UQDB}\\
		\hline
		\multicolumn{2}{|c|}{Metrics-Judge} & \multicolumn{2}{|c|}{LLM-judge} & \multicolumn{2}{|c|}{Combo-Eval}&\multicolumn{2}{c|}{Metrics-Judge} & \multicolumn{2}{|c|}{LLM-judge} & \multicolumn{2}{|c|}{Combo-Eval}  \\
		\hline
        	\multicolumn{2}{|c|}{69.73} & \multicolumn{2}{|c|}{$80.72\pm$0.8} & \multicolumn{2}{|c|}{$80.88\pm$0.4} & \multicolumn{2}{|c|}{68.06} & \multicolumn{2}{c|}{$75.82\pm$0.4} & \multicolumn{2}{|c|}{$76.98\pm$0.4}\\
		\hline
            \hline        
		C0 & C1 & C0 & C1 & C0 & C1 & C0 & C1 & C0 & C1 & C0 & C1  \\
		\hline
		\begin{tabular}{@{}c@{}}53.27 \\ $\pm$0\end{tabular} & \begin{tabular}{@{}c@{}}86.19 \\ $\pm$0\end{tabular} & \begin{tabular}{@{}c@{}}70.79 \\ $\pm$1.1\end{tabular} & \begin{tabular}{@{}c@{}}90.65 \\ $\pm$0.4\end{tabular} & \begin{tabular}{@{}c@{}}73.22 \\ $\pm$0.7\end{tabular} & \begin{tabular}{@{}c@{}}88.55 \\ $\pm$0.2\end{tabular} & \begin{tabular}{@{}c@{}}52.05 \\ $\pm$0\end{tabular} & \begin{tabular}{@{}c@{}}84.06 \\ $\pm$0\end{tabular} & \begin{tabular}{@{}c@{}}61.77 \\ $\pm$0.6\end{tabular} & \begin{tabular}{@{}c@{}}89.87 \\ $\pm$0.3\end{tabular} &\begin{tabular}{@{}c@{}}64.75 \\ $\pm$0.5\end{tabular} & \begin{tabular}{@{}c@{}}89.22 \\ $\pm$0.3\end{tabular} \\
    \hline
	\end{tabular}
		\caption{F1 average (macro) across classes and F1 score per class, with standard deviation across 10 runs, for evaluation methods - metrics (ROUGE1-recall with threshold to determine decision), LLM-as-a-judge, and Combo-Eval method, across GT and UQDB scenarios. class 0 (C0);  class 1 (C1). Judge LLM is GPT-4o.}
		\label{tab:eval}
	\end{center}
\end{table*}

\paragraph{LLM-as-a-Judge:}
Table~\ref{tab:eval} reveals that LLM-based judgment outperforms metric-based methods in both GT and UQDB scenarios. Notably, performance is more robust for class 1, though identifying class 0 instances remains more challenging.

\begin{figure}
    \centering
    \includegraphics[width=1.0\linewidth]{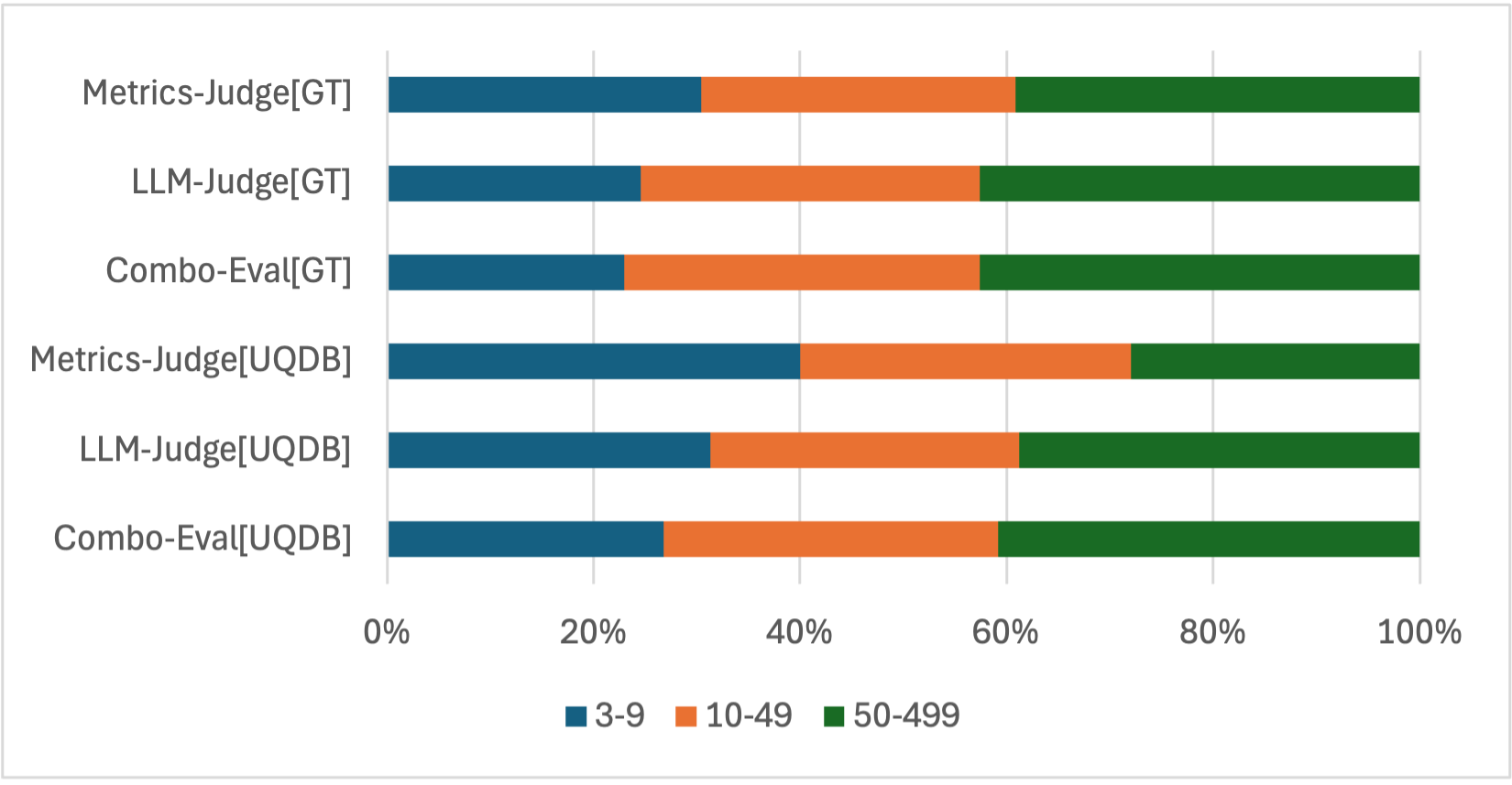}
    \caption{Breakdown of incorrect judgments by result size across evaluation methods (Metrics-based, LLM-judge, and Combo-Eval) for GT and UQDB scenarios, showing higher misjudgments by LLM-as-a-judge on higher result sizes.}
    \label{fig:misjudgement-resultsize}
\end{figure}

Dissecting false judgments by result size, both UQDB and GT show higher evaluation inaccuracies with larger sizes, as seen in Figure~\ref{fig:misjudgement-resultsize}. Although LLM-judge yields fewer incorrect judgments than Metrics-judge, it exhibits a higher proportion of false judgments at larger result sizes, as detailed in Appendix~\ref{sec:misjudge}. 
This shows that at larger result sizes, not only is it harder for LLMs to generate correct NLRs (Table~\ref{tab:llmeval}), but it is also more difficult for LLM-based judges to accurately evaluate model-generated NLRs.

\paragraph{Combo-Eval:}
Several experiments were conducted including combining metrics and LLM-judge output as features and training a machine learning model which didn't yield a robust model and the overall performance did not improve. We also tried injecting extra knowledge of the metrics into the prompt or LLM-as-a-judge method which did not improve the performance either. 

\begin{figure*}[t]
    \includegraphics[scale=0.6]{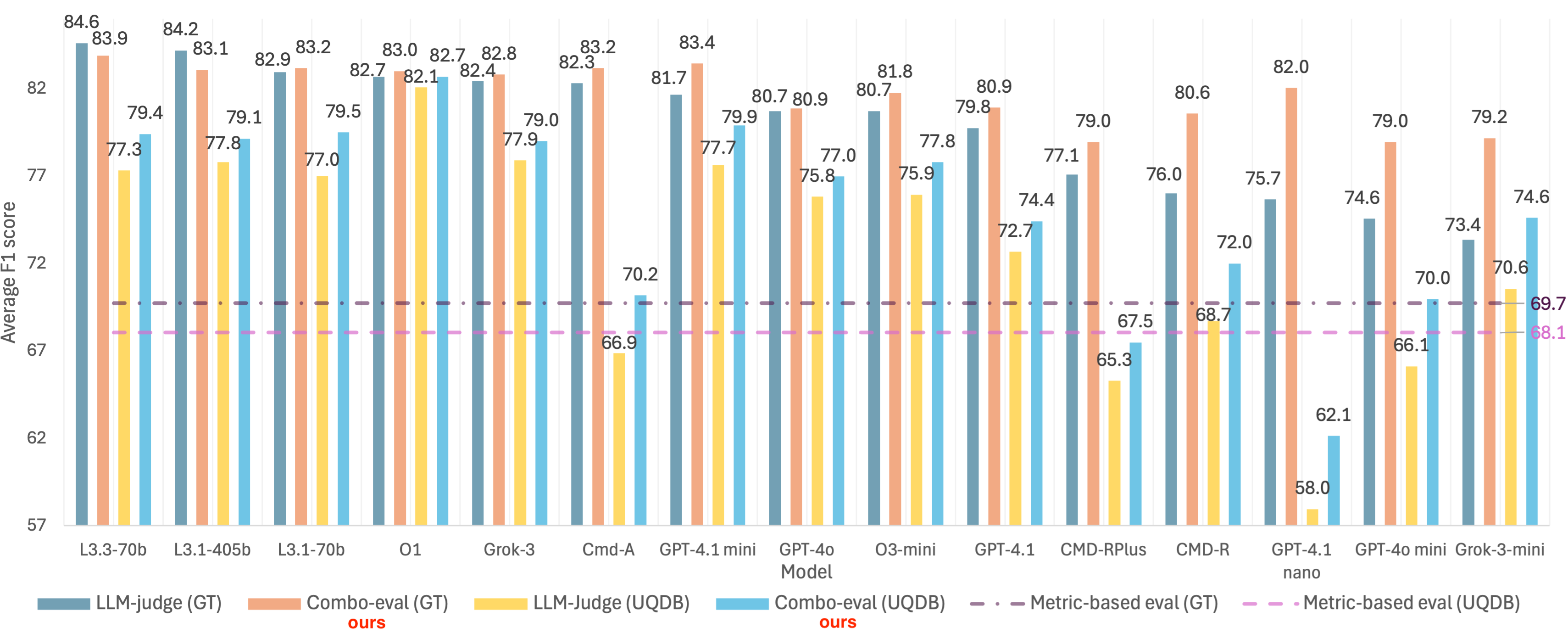}
    \caption{Macro F1 scores across judge LLMs and evaluation methods for GT and UQDB scenarios. This figure shows that Combo-Eval surpasses LLM-judge across different judge LLMs and highlights trends across judge models. LLMs are ordered by their LLM-judge scores on GT, highest to lowest. Results in tabular format are in Appendix~\ref{sec:judges}}
    \label{fig:evaljudgelms}
\end{figure*}

Using a single threshold on ROUGE-recall does not suffice, but the difference between class 1 and class 0 tends to be more informative on the extreme values (more information is shared in Appendix~\ref{sec:Thresholds}). Combo-Eval effectively transcends limitations associated with single thresholds by dynamically applying metric thresholds and subsequent LLM evaluation. This method demonstrated superior performance, validating its utility across varied judge models, as demonstrated in Figure~\ref{fig:evaljudgelms}. 

For our test set, \textbf{\textit{Combo-Eval reduced the LLM calls by 61.43\% in GT and 24.48\% for UQDB}}. This method optimizes the evaluation by passing data through a low-complexity calculation first, and passing only a subset of data through an LLM. 

\paragraph{Combo-Eval across Models:} As seen in Figure~\ref{fig:evaljudgelms}, small judge models (e.g., Grok-3-mini, GPT-4.1 nano, GPT-4o mini) exhibit a prominent improvement with the Combo-Eval framework, demonstrating effective performance at reduced computational costs. Larger models maintain smaller benefits, underscoring Combo-Eval as a strategic choice to balance accuracy and resource consumption. This pattern is seen in both GT and UQDB scenarios. Generally, Combo-Eval outperforms, signifying its robustness in varied scenarios. Also, Combo-Eval's reduced standard deviation in results (Table~\ref{tab:eval}) stems from fewer samples requiring LLM evaluation, stabilizing evaluation consistency. 

\paragraph{GT vs. UQDB:} Unlike GT, UQDB isn’t the ground truth NLR, but the source information that makes up the contents of the NLR. Thus, we see models with stronger reasoning capabilities, such as O1, exhibit consistent performance across GT and UQDB scenarios, doing better in the UQDB scenario compared to other models. On the contrary, most other models do better with GT than UQDB by an average of 7.27\% for LLM-judge method and 6.71\% for Combo-Eval (Appendix~\ref{sec:judges}). However, with GT references available, newer Llama models and Grok-3 are formidable contenders, indicating various models' adaptability.

Consistently, GT-based evaluations outperform UQDB, though UQDB offers a viable alternative when GT is inaccessible. UQDB results are much worse for Cmd-A, Cmd-R+, Cmd-R, GPT-4.1 nano, and GPT-4o mini, some even underperforming compared to metrics-judge, indicating potential limitation in inherent reasoning capabilities compared to other models evaluated.

\paragraph{Model Ranking for NLR Generation:} Appendix~\ref{sec:genllms} elucidates rankings among NLR generation models using the three evaluation methods. Our evaluations establish that irrespective of methods, LLM rankings remain consistent, affirming metrics as practical ranking tools for model performance analysis.

\paragraph{Temperature Influence:} While low temperature settings are known to keep LLMs judgment closer to the factual content \citep{lowtempeval, salinas2025tuning}, our experiments show that a judge model's temperature parameter change only minimally enhanced or deteriorated the judgment in different directions for different judge models, showing lack of any notable impact or trend. Combo-Eval had similar improvements over LLM-Judge across temperatures.
The trend between LLM-judge and Combo-Eval remained unchanged with temperature changes. 
Experiment details and results are in Appendix~\ref{sec:tempjudge}.

\paragraph{Extension and Future Directions:} 
This work lays the foundation for systematic evaluation of existing and emerging LLMs on NLR generation and judgment tasks. Future studies can leverage the benchmark as a new dimension for model comparison and assessment under controlled conditions.

Beyond Text-to-SQL systems, the framework can be extended to other settings that require narrating structured or tabular data. For instance, NLRs can support schema enrichment, expanding database metadata with descriptive language that helps users understand available information and formulate better queries. In interactive systems, such narrations can bridge the gap between user intent and schema representation (e.g., mapping "men's department" to the canonical entity MENS\_DEPT), enhancing question understanding.

Another future direction includes comparing Combo-Eval with hard metrics in reasoning-aware datasets such as SciGen \citep{moosavi2021scigen}, which feature expert-annotated table descriptions requiring arithmetic reasoning.

Insights from our benchmarking results open up opportunities to improve LLM capabilities. Future work may explore improved LLM training strategies or data augmentation methods to improve models' ability to handle large result sets and NLR generation performance.

\section{Conclusion}
Using NL to communicate with DBs is a valuable industry application. 
Our study identifies incomplete information as a predominant source of error in NLRs, with model performance deteriorating as result size increases. We introduce Combo-Eval method, a new evaluation framework that aligns closely with human judgment of NLR correctness compared to traditional metrics-based and LLM-based methods. Notably, Combo-Eval achieves this alignment while significantly reducing computational demands by minimizing LLM calls, making it especially advantageous in large-scale or resource-constrained environments. Combo-Eval's differentiated improvements are particularly evident when using smaller judge models, offering a cost-effective solution without compromising accuracy. Our contribution also includes the release of the NLR-BIRD dataset, providing a valuable resource for benchmarking and further research in this domain. Our experiments demonstrate that the presence of ground truth data enhances evaluation accuracy across methods. In scenarios where ground truth is unavailable, UQDB proves to be a viable alternative.

\section*{Limitations}

Our study, while advancing NLR evaluation methodologies, does present certain limitations. While the NLR-BIRD dataset provides a robust foundation, its domain coverage may not fully capture the diversity of all industrial applications. These aspects highlight the need for continual dataset expansion and further exploration of context-specific evaluation strategies. Additionally, improving NLR generation by LLMs is not the paper’s focus. Future LLMs could benefit from training in these areas to enhance their table narration abilities and reduce information loss— the number one reason behind incorrect NLRs as uncovered by our work. Emphasizing completeness in data-based narration can inherently improve LLM performance in this task. 
Moreover, our analysis shows varied performance based on result-set size which may inform potential next steps in this area. For large result-sets, there is a significant gap in performance, indicating that summarization or rule-based approaches might be needed to ensure NLR completeness.

\bibliography{custom}

\appendix
\section{Appendix}
\label{sec:appendix}

\subsection{Related Work and Background}
\label{sec:relatedwork}

Large language models (LLMs) have become foundational tools in natural language processing (NLP), a field that enables diverse domain analysis across modalities such as speech \citep{speechpaper, singh2022} and text. 
LLMs have rapidly transformed real-world applications such as visual question answering \citep{pattnayak2024survey,pattnayak2025hybrid}, product or content search and analysis \citep{meghwani-etal-2025-hard, icxpatent, singh2021}, interactive assistants, document understanding,  \citep{patel2024llm,patel2025sweeval,agarwal2025aligningllmsmultilingualconsistency,agarwal-etal-2025-fs}, accessibility for visually impaired users \citep{panda2025accessevalbenchmarkingdisabilitybias,panda2025whosaskinginvestigatingbias,panda2025daiqauditingdemographicattribute}, and synthetic data-pipelines \citep{dua-etal-2025-speechweave,agarwal2024techniques,agarwal2024synthetic}.
Structured data understanding and connecting language interfaces to structured data has gained more momentum in the last few years.

Traditionally in published studies on natural language interfaces to DBs, the focus has been either on converting text to SQL \citep{iacob-etal-2020-neural} or long-form table QA which stands for trying to extract answers present in tables directly, without using SQL. 

Surrounding text-to-SQL-based applications, a lot of work has been presented such as question/query decomposition for breaking down the query into sub-queries and combining the results \citep{Pourreza2024-nn}, SQL correction for self-checking syntactical issues in the generated query \citep{Chen2023, Chen2023TexttoSQLEC}, producing SQL in the presence large schema and multi-table retrievals \citep{chen-etal-2024-table, https://doi.org/10.48550/arxiv.2407.01183}, schema linking \citep{https://doi.org/10.48550/arxiv.2405.16755}, noise in SQL generation \citep{wretblad-etal-2024-understanding}, SQL generation systems \citep{tian-etal-2023-interactive}, scaling \citep{wang2025dbcopilotnaturallanguagequerying} and more to get better SQL generations from models \citep{https://doi.org/10.48550/arxiv.2307.07306, yang-etal-2024-synthesizing, sun-etal-2023-sqlprompt, Li2024}. However, this primarily targets complexities around the SQL generation part and the focus on industry applications consuming the results of such a system has been missing. In conversational systems with Text-to-SQL backing, the user asks a question in NL and also expects the response back in NL rather than structured table or JSON type of representation. 
One of the mentions of NLR generation is in \citep{https://doi.org/10.48550/arxiv.2408.14717}, where the work briefly touched upon generating responses from database results in the context of questions requiring a combination of database schema and real-world knowledge. However, there have been no mentions of evaluation of these NL responses. Thus, we find that there has been limited work done on the natural language representations of the result-set fetched after the produced SQL is run against the DB. Existing work \citep{https://doi.org/10.48550/arxiv.2408.14717} uses language models to handle such conversions and no study presents how well language models are able to represent such data in NL, nor any evaluation methods have been discussed for this specific problem. Errors/information loss induced by this NLR generation step is not studied.

There has also been work in extracting NL answers from tables directly, specifically around short-form and long-form Table QA (LFTQA) \citep{Eratta} \citep{zhao-etal-2024-tapera} \citep{nan-etal-2022-fetaqa}. Most of these tasks have been around answering questions where the answer to the question needs to be looked up in tables. In essence, partial information from the table forms the final answer, combined with other general/analytical reasoning in some of the benchmarks. 
On the contrary, our work is focussed on the task of narrating a tabular answer to a user question in natural language. The task is not looking for answers in the table or relying on reasoning from LLMs to fetch answers from tables, but in majority of the questions, the full table provided in the actual answer and the job of the model is to correctly describe the information in the table in natural language given the user question's context. In particular, the benchmark we provide is applicable to Text-to-SQL applications in the real world including conversational systems and are different from LFTQA.

The work presented in \citep{parikh-etal-2020-totto} focuses on obtaining descriptions from highlighted sections of tables. It proposes a controlled generation task where, given a Wikipedia table and highlighted cells, the goal is to generate a descriptive sentence. However, this dataset does not explicitly represent the verbalization of tables in a Text-to-SQL application, and there is currently no benchmark available for this specific task.

While evaluation methods for NLR generation from tabular result-set have not been explored, there has been work in evaluating responses extracted from tabular sources in long-form Table QA. We draw from LFTQA research on evaluation \citep{wang-etal-2024-revisiting} where it shows that existing automatic metrics fail in assessing the answers generated by LLM-based systems (e.g.,BLEU, ROUGE, METEOR, and TAPAS-Acc). LLM-based metrics demonstrate a significant improvement over traditional automated metrics in terms of correlation with human evaluation. 
Even though the task we are discussing here is not LFTQA, it is related to tables and thus we can derive some learning from the field. We evaluate metrics and LLM based evaluation methods for NLRs as well. We use similar criteria for evaluating NLRs using LLM-as-a-judge for faithfulness and comprehensiveness. We also take it forward and present a Combo-Eval method that uses both metrics and LLM-judge, and surpasses LLM-judge's performance in addition to reducing number of LLM calls required.

If LLMs can handle long-form table question answering, a task that seems more complex than converting tables to natural language, one might assume that the latter would be a relatively simple task lacking challenges. However, this is not the case. Even long-form question answering on tables remains an unsolved problem, posing numerous challenges. particularly in industrial settings, where accuracy is paramount, limiting its industry adoption. Our study reveals that even for table-to-NL tasks, LLMs struggle to perform effectively. In this paper, our prime focus is on evaluation of these NLRs generated from tables across evaluation techniques and models.
\subsection{Data Curation Process}
\label{sec:datacuration}
The NLR-BIRD dataset was developed by extracting and processing SQL queries from the BIRD-dev dataset using SQLite. These queries were executed against the database to retrieve the result sets, serving as the basis for creating natural language responses (NLRs).

Labeling involved contributions from four professionals with backgrounds in database management, data science, engineering, and related domains. To streamline the annotation process, a subset of queries was initially processed by various LLMs to create a baseline. Annotators improved upon these by either refining LLM-generated responses or crafting new NLRs from scratch.

For quality assurance, each annotator began with a small set of random samples, both unlabeled and baseline-labeled, to calibrate their understanding of the task. Through iterative discussions, annotators developed shared criteria for assessing NLR correctness: 

- An NLR is considered correct if it comprehensive and encapsulates all pertinent information from the result-set relevant to the user's question.

- A correct NLR should exhibit faithfulness. Inclusion of excessive, unverifiable information renders an NLR incorrect.

- A correct NLR should be readable. While a consistent format is encouraged, variations are permissible if readability and completeness are maintained.

Example of handling ambiguous queries include interpreting “Which constructors have been ranked 1?” where the SQL-derived DB results list multiple constructors. Annotators were instructed to treat all listed constructors as sharing rank 1, consistent with the ground-truth logic used to generate the DB results.

A post-annotation review process involved multiple reviewers to cross-check agreements, achieving an 83.3\% agreement rate among annotators.

\subsection{Dataset Attributes}
\label{sec:datattrs}

\begin{figure}[t]
	\includegraphics[width=\columnwidth]{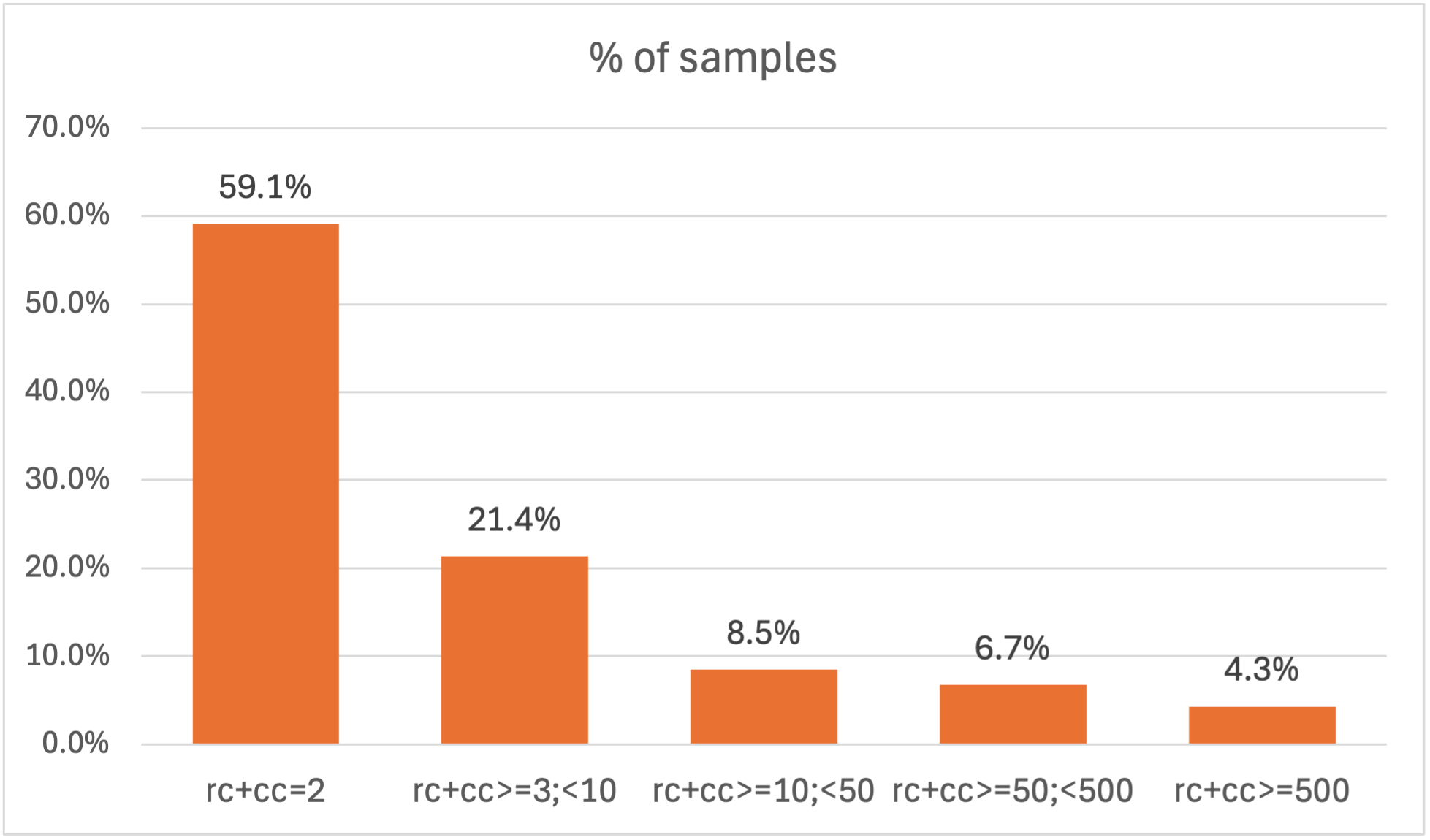}
	\caption{Result size distribution in the BIRD-dev dataset. rc=row count; cc=column count.}
	\label{fig:dataset_result_analysis}
\end{figure}

Figure~\ref{fig:dataset_result_analysis} shows the result size disctribution in the BIRD-dev dataset. The NLR-BIRD datasets contain NLRs for all result sizes rc+cc<500.

In the NLR-BIRD dataset, he maximum word length is 92 characters, typically representing URLs. NLRs are composed of 19,601 numeric characters and 62,140 alphabetic characters.

\begin{table}
	\begin{center}
		\begin{tabular}{|p{1.15cm}||p{0.75cm}|p{0.75cm}|p{0.75cm}|p{0.75cm}|p{0.75cm}|}
			\hline
			& Total & rc+cc = 2 & 3<= rc+cc <10 & 10<= rc+cc <50 & 50<= rc+cc <500\\
			\hline
			\hline
			count & 1468 & 907 & 328 & 130 & 103 \\
			mean(c) & 311 & 74  & 120  & 434 & 2853 \\
			mean(w) & 49 & 13 & 20 & 65 & 444 \\
			std(c) & 1421 & 62 & 112 & 659 & 4611 \\
			std(w) & 236 & 10 & 19 & 100 & 784 \\
			min(c) & 3 & 3 & 8 & 40 & 13 \\
			min(w) & 1 & 1 & 2 & 5 & 5 \\
			25\%(c) & 57 & 51 & 68 & 125 & 636 \\
			25\%(w) & 9 & 8 & 12 & 22 & 114 \\
			50\%(c) & 81 & 68& 103 & 244 & 1470 \\
			50\%(w) & 14 & 12& 18 & 40 & 202 \\
			75\%(c) & 123 & 89  & 142 & 442 & 3266 \\
			75\%(w) & 21 & 16 & 23& 67 & 449 \\
			max(c) & 27892 & 1481 & 1508 & 6014 & 27892 \\
			max(w) &  4931 &  237 & 252 & 926 & 4931 \\
			\hline
		\end{tabular}
		\caption{NLR statistics: Character (c) and word (w) counts across result sizes (rc+cc) and overall.}
		\label{tab:nlrstats}
	\end{center}
\end{table}

Table~\ref{tab:nlrstats} illustrates the distribution of character and word counts for NLRs across varied result-set sizes.

\begin{figure}[t]
	\includegraphics[width=\columnwidth]{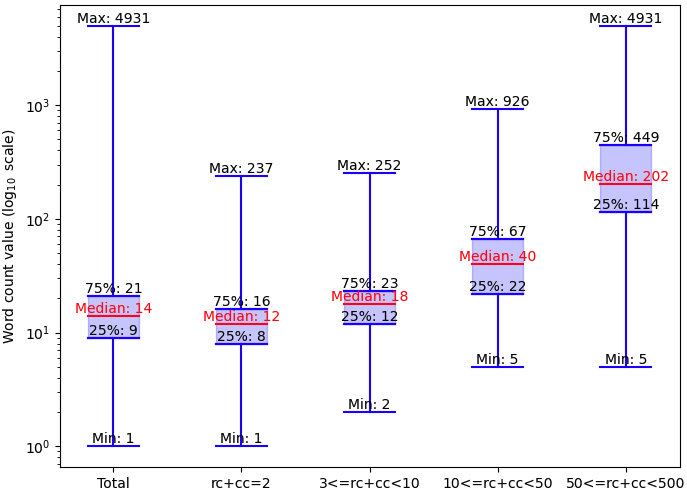}
	\caption{Box plot for word count distribution in NLRs across various result-set sizes.}
	\label{fig:boxplot-words}
\end{figure}

\begin{figure}
	\includegraphics[width=\columnwidth]{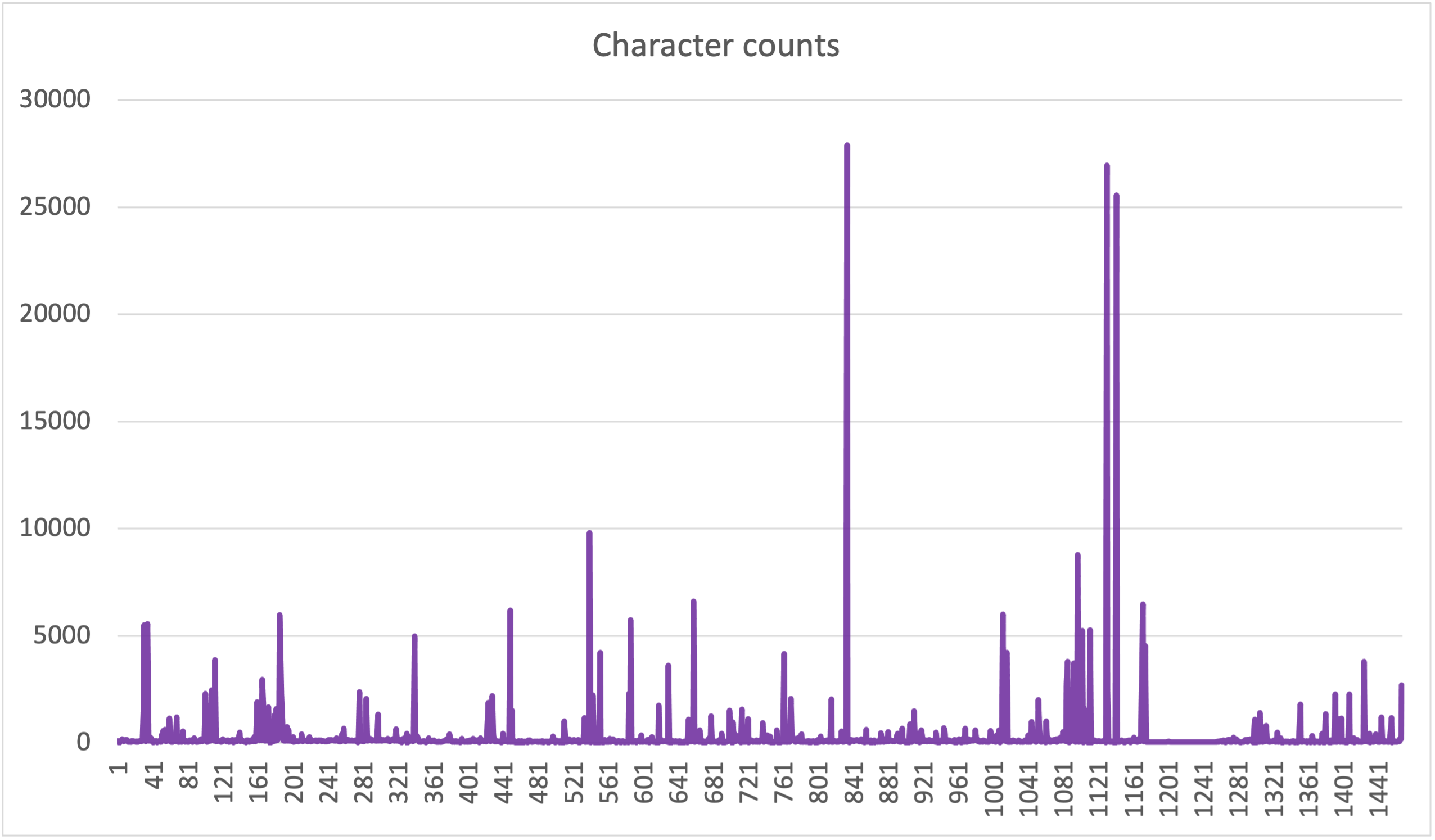}
	\caption{Character count distribution in NLRs of the NLR-BIRD dataset. (x-axis: sample number, y-axis: character count.)}
	\label{fig:nlr_size_distn}
\end{figure}

Figure~\ref{fig:boxplot-words} and Figure~\ref{fig:nlr_size_distn} depict the word count distribution and character count variation among NLRs, respectively.

\subsection{Prompts}
\label{sec:prompts}
\textbf{Prompt for NLR generation:}

\begin{lstlisting}[language=Python]
tbl_str = table.to_json(orient="records", lines=True) # table is a pandas dataframe

prompt = """For the following question, use the Answer provided to generate a response in plain text. Do not make up any information, only use what can be found in the table to return the plain text response. Do not compute any trend. Do not calculate any numbers. Do not miss any answer row.


Question: {question}

Answer: {tbl_str}

Response:"""
\end{lstlisting}

\textbf{LLM judge prompt:}

\begin{lstlisting}[language=Python]
judge_prompt = """Question: {q}

Actual answer:{ip}

Model generated answer:{op}

For the above question, you have the correct answer (`Actual answer`) known and a model generated answer. Compare if the model generated answer contains complete and correct information to answer the question as the actual answer. 
It is ok if the format is different but core information in the model generated answer should match the correct answer as it relates to the question. 
Does the model generated answer hold the same information as the actual answer? Say True or False.

Evaluation:"""
\end{lstlisting}
The value for ip above is the ground truth NLRs for GT and the user question + db results for UQDB.

\subsection{Inference Parameter Settings}
\label{sec:params}
The following settings were used where applicable.
\begin{itemize}
    \item max\_new\_tokens: 2000
    \item temperature: 0.01
    \item top\_p: 0.95
    \item top\_k: 10
    \item frequency\_penalty: 1.1
\end{itemize}

\subsection{Metrics}
\label{sec:evalsmetrics}
Table~\ref{tab:metric-eval-all} contains the different metrics considered for metric-based evaluation, along with a threshold for each to determine decision boundary and results representing alignment with human assessment. The thresholds were determined on the dev set and the results shared below reflect the results of the thresholds applied on the test set.

\begin{table*}
	\begin{center}
		\begin{tabular}{|l|l|c|c|c||c|c|c|}
        \hline
			 &  & Recall & Prec & F1 & Recall & Prec & F1\\
          
             Metric & Thresholds & \multicolumn{3}{c||}{GT} & \multicolumn{3}{c|}{UQDB}\\
			\hline
			Cosine similarity & GT=0.7 UQDB=0.65 &	0.65 & 0.62 & 0.63 & 0.61 & 0.6 & 0.6 \\
			BERTscore recall & GT=0.3 UQDB=0.3 &	0.54 & 0.69 & 0.6 & 0.59 & 0.77 & 0.67 \\
			BERTscore precision & GT=0.7 UQDB=0.6 &	0.65 & 0.67 & 0.66 & 0.55 & 0.63 & 0.59 \\
			BERTscore F1 & GT=0.45 UQDB=0.41 &	0.56 & 0.66 & 0.6 & 0.56 & 0.68 & 0.62 \\
			ROUGE-1 recall & GT=0.45 UQDB=0.4 &	0.68 & 0.71 & 0.7 & 0.68 & 0.68 & 0.68 \\
			ROUGE-1 precision & GT=0.8 UQDB=0.65 &	0.63 & 0.62 & 0.63 & 0.56 & 0.58 & 0.57 \\
			ROUGE-1 F1 & GT=0.7 UQDB=0.5 &	0.72 & 0.69 & 0.71 & 0.69 & 0.71 & 0.7 \\
			ROUGE-2 recall & GT=0.4 UQDB=0.3 &	0.72 & 0.7 & 0.71 & 0.68 & 0.66 & 0.67 \\
			ROUGE-2 precision & GT=0.7 UQDB=0.43 &	0.65 & 0.62 & 0.64 & 0.58 & 0.58 & 0.58 \\
			ROUGE-2 F1 & GT=0.45 UQDB=0.43 &	0.71 & 0.68 & 0.69 & 0.67 & 0.64 & 0.66 \\ 
			ROUGE-L recall & GT=0.45 UQDB=0.4 &	0.69 & 0.71 & 0.7 & 0.67 & 0.67 & 0.67 \\
			ROUGE-L precision & GT=0.8 UQDB=0.42 &	0.63 & 0.61 & 0.62 & 0.54 & 0.61 & 0.57 \\
			ROUGE-L F1 & GT=0.7 UQDB=0.5 &	0.71 & 0.66 & 0.69 & 0.7 & 0.67 & 0.68 \\
			\hline
		\end{tabular}
		\caption{Decision boundary thresholds (determined using the dev set) across various metrics, along with results on test set (macro average recall, precision, and F1 scores) indicating alignment with human assessments of NLR correctness. Thresholding on ROUGE scores aligns more closely with human assessments than cosine similarity and BERT scores.}
		\label{tab:metric-eval-all}
	\end{center}
\end{table*}

\begin{table*}
	\begin{center}
		\begin{tabular}{|l|c|c|c|c|}
			\hline
			& \multicolumn{4}{|c|}{Median of the metric computed between}\\
			& \multicolumn{4}{c|}{model generated NLR and}\\

			Metric & \multicolumn{2}{|c|}{\phantom{x}\hspace{4ex}GT\hspace{8ex}\phantom{x}} & \multicolumn{2}{|c|}{UQDB} \\
			\hline
			 & \phantom{x}\hspace{1ex}class 0\hspace{1ex} & \phantom{x}\hspace{1ex}class 1\hspace{1ex} & \phantom{x}\hspace{1ex}class 0\hspace{1ex} & \phantom{x}\hspace{1ex}class 1\hspace{1ex} \\
			\hline
			Cosine similarity & 0.58 &	0.75 & 0.64 & 0.72 \\
			BERTscore recall & 0.70 & 0.79 & 0.51 & 0.52\\ 
			BERTscore precision & 0.74 & 0.82 & 0.68 & 0.73\\ 
			BERTscore F1 & 0.70 & 0.81 & 0.58 & 0.60\\ 
			ROUGE-1 recall & 0.47 & 0.86 & 0.39 & 0.65\\
			ROUGE-1 precision & 0.78 & 0.93 & 0.81 & 0.92\\
			ROUGE-1 F1 & 0.53 & 0.83 & 0.49 & 0.69\\
			ROUGE-2 recall & 0.27 & 0.69 & 0.22 & 0.4\\ 
			ROUGE-2 precision & 0.48 & 0.79 & 0.5 & 0.67\\ 
			ROUGE-2 F1 & 0.31 & 0.67 & 0.24 & 0.49\\ 
			ROUGE-L recall & 0.44 & 0.81 & 0.36 & 0.58\\ 
			ROUGE-L precision & 0.71 & 0.91 & 0.73 & 0.83\\ 
			ROUGE-L F1 & 0.48 & 0.76 & 0.42 & 0.63\\
			\hline
		\end{tabular}
		\caption{Median values broken down by human evaluation category (0-NLRs deemed incorrect by human; 1-NLRs deemed correct by human) across scores (cosine similarity, BERTscore, and ROUGE) representing computation of model produced NLRs wrt GT NLRs as well as wrt UQDB.}
		\label{tab:appendix-metrics}
	\end{center}
\end{table*}

We calculated ROUGE-1, ROUGE-2, ROUGE-L, BERTscore, and cosine similarity for the model generated NLRs (compared to both GT and UQDB) and computed their medians per class Table~\ref{tab:appendix-metrics} shows the results.

\subsection{Thresholds Used for Experiments}
\label{sec:Thresholds}

\textbf{Metrics:}

A threshold was determined on the dev set to make the decision boundary using grid search. This was done while optimizing for macro F1 score. We searched for this threshold for ROUGE1-recall metric, which was choosen specifically for its simplicity, general applicability on different types of text, and large difference in median scores between class 0 and class 1.

Threshold of 0.45 for GT and 0.4 for UQDB yielded best macro F1 score (across class 0 and class 1) on the dev set. The Results section shared the results from application of these thresholds on the test set.

\textbf{Combo-Eval:}

\begin{figure}[h]
\includegraphics[width=\columnwidth]{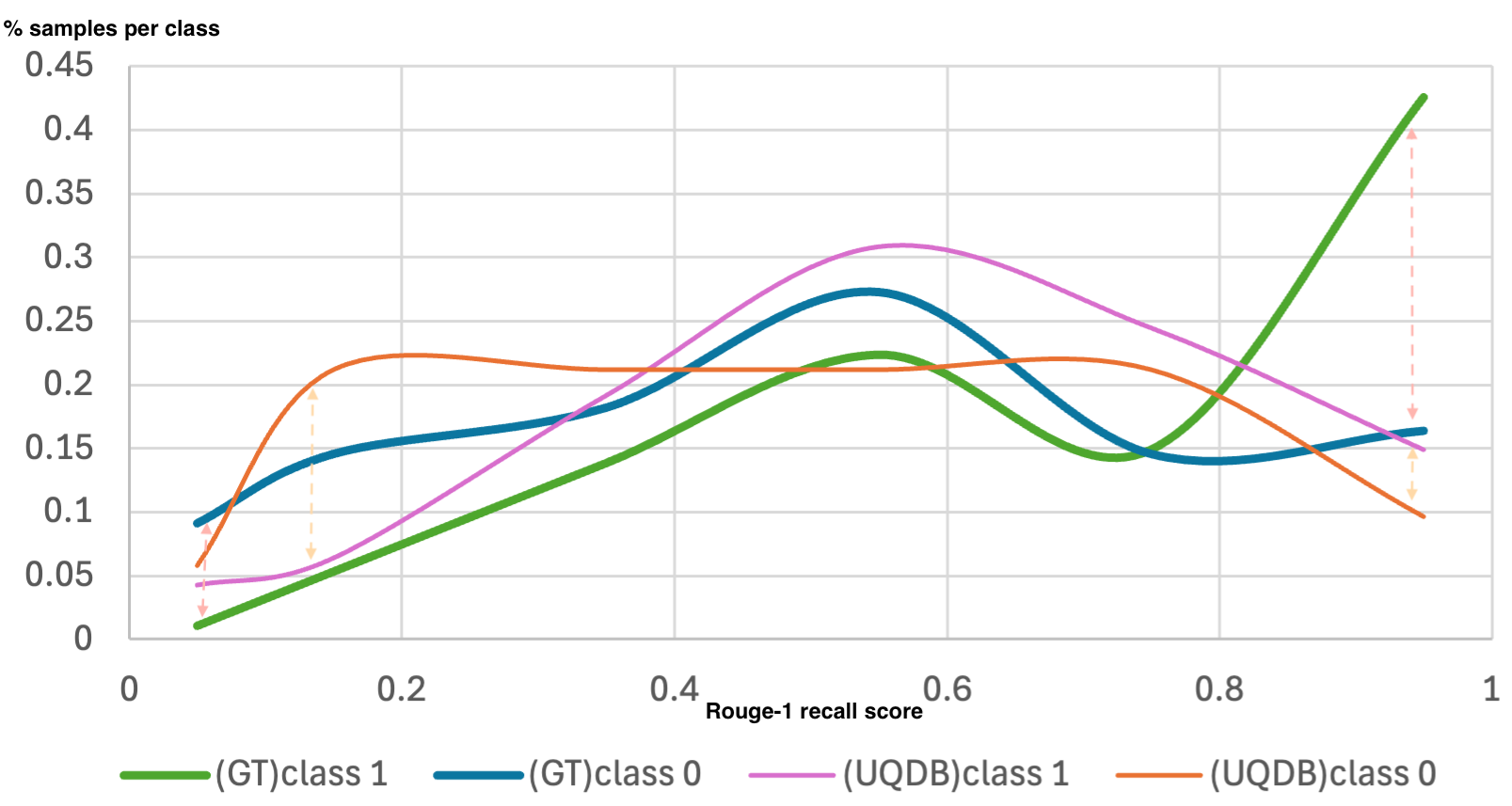}
	\caption{Trend for \% of samples with different ROUGE 1 recall scores for class 1 and for class 0 for GT vs. model outputs and UQDB vs. model outputs. The plot shows the extreme scores tend to show more difference between class 1 and class 0, indicating the potential for choosing thresholds that are more likely to contain true positives and true negatives, and minimizing false positives and false negatives.}
\label{fig:scatter}
\end{figure}

As shown in Figure~\ref{fig:scatter}, the extreme scores provide a better indication of the accuracy of an NLR. The difference between class 1 and class 0 ROUGE recall scores tends to be more informative on the extreme values, where the difference between \% of samples from class 1 and class 0 for the score threshold is not only high but also the class with lower number is samples has a very small presence on those scores.

After conducting a grid search on the lower and upper extremes of ROUGE-1 recall scores using the development set, we established thresholds for both ends. This grid search was optimized for macro F1 score to best align with human assessments of NLRs, where F1 scores were averaged for class 0 (NLRs deemed incorrect by humans) and class 1 (NLRs deemed correct by humans).

The following thresholds were used.

\(th_0l = 0\) for GT; \(0.05\) for UQDB

\(th_0u = 0.1\) for GT; \(0.1\) for UQDB

\(th_1l = 0.9\) for GT; \(0.87\) for UQDB

\(th_1u = 1\) for GT and UQDB

These thresholds were then applied to test set and that represents the numbers shared in the Results section~\ref{sec:resultsanddis}.

\subsection{LLM Generation Errors}
\label{sec:err}

\begin{figure}[t]
\begin{center}
	\includegraphics[width=\columnwidth]{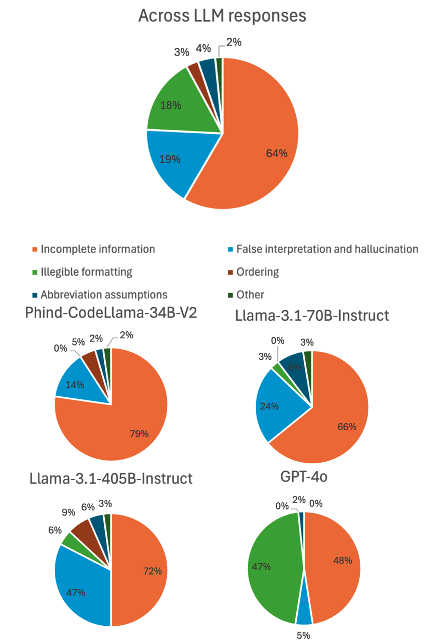}
	\caption{Reasons for inaccurate NLRs - overall and by LLM.}
	\label{fig:faults}
\end{center}
\end{figure}

Some errors produced by LLM generated NLRs have common grounds for what went wrong. Figure~\ref{fig:faults} shows the trends for breakdown of the error types. 

Clarifying the source of errors, all identified inaccuracies stem from LLM-generation of NLR, as SQL execution was based on ground-truth SQLs provided in the BIRD dataset, assumed to represent correct data.

Examples of correct and incorrect NLRs by generation models are shared below.
\begin{itemize}
    \item{\textbf{Model}: Phind-CodeLLma-34B-v2

    \textit{Correct NLR:}
    
    Q: "In which city can you find the school in the state of California with the lowest latitude coordinates and what is its lowest grade? Indicate the school name.

NLR: "The school with the lowest latitude coordinates in the state of California is Central Middle - RISE located in Oroville. This school offers classes from grade 7 onwards."

\textit{Incorrect NLR:}

Q: "What are the webpages for the Los Angeles County school that has between 2,000 and 3,000 test takers?"

NLR: "The webpage for the Los Angeles County school with between 2,000 and 3,000 test takers is www.avdistrict.org."

NLR is missing information. The DB result set returned websites 'www.avdistrict.org', 'www.lbschools.net', and 'www.hartdistrict.org'.
}
    \item{\textbf{Model}: Llama3.1-70B-Instruct 

    \textit{Correct NLR:}

Q: "Give the code of sets have expansion commander type?"

NLR: "The sets with expansion commander type are C13, C14, C15, C16, C17, CM1, and CMD.

\textit{Incorrect NLR:}

Q: "When was the set of cards with "Ancestor's Chosen" released?"

NLR: "The set of cards with "Ancestor's Chosen" was released on 2002-05-27."

This is incorrect because the DB results returned three dates (['2007-07-13'], ['2002-05-27'], ['2018-12-07']), but the LLM assumed only one is correct. 

    }
    \item{
    \textbf{Model}: Llama-3.1-405B-Instruct

\textit{Correct NLR:} 

Q: "List the football players with a birthyear of 1970 and a birthmonth of October."

NLR: "Chris Swailes, Dean Kiely, Edwin van der Sar, and Sander Boschker"

\textit{Incorrect NLR:} 

Q: "Which set is not available outside of the United States and has foil cards with Japanese writing on them? Please include the set ID in your response."

NLR: "Commander's Arsenal (50) is not available outside of the United States and has foil cards with Japanese writing on them."

Names and IDs returned included ["Commander's Arsenal", '50'], ['Premium Deck Series: Fire and Lightning', '307'], ['From the Vault: Exiled', '522'], ['From the Vault: Relics', '523'], ['From the Vault: Legends', '524'], ['From the Vault: Realms', '525'], ['From the Vault: Twenty', '526'], ['From the Vault: Annihilation', '527'], ['From the Vault: Angels', '528'], ['From the Vault: Lore', '529']. The NLR contained incomplete information.
}
\item{
\textbf{Model}: GPT4o

\textit{Correct NLR:} 

Q: "What is the sum that client number 4's account has following transaction 851? Who owns this account, a man or a woman?"

NLR:"The sum that client number 4's account has following transaction 851 is 15096. The owner of this account is a man."

\textit{Incorrect NLR:}

Q: "What is the amount of debt that client number 992 has, and how is this client doing with payments?"

NLR: "Client number 992 has a debt amount of 53076 and is doing well with payments, as indicated by the status "A"." 

Assumption for the meaning of A is considered incorrect.
For another example of a bad NLR for this model, see Appendix~\ref{sec:gptformateg}.
    }
\end{itemize}

\subsection{Format Inconsistency Error Example}
\label{sec:gptformateg}
While manually reviewing LLM generation of NLRs, we noticed format inconsistencies impacting readability of NLRs, specially in high volumes for GPT-4o as the generation model. An example of such an NLR is as follows.

User question: Provide the IDs and age of the client with high level credit card, which is eligible for loans.

NLR generated based on DB results (21 rows) is shared below. For readability and space, some lines were truncated (represented by symbol <...TRUNC>) where the same format pattern was repeating.

\begin{quote}
\small
Account ID: 66, District Name: Rychnov nad Kneznou, District Region: east Bohemia\\
\\
Account ID: 273, District Name: Karlovy Vary, District Region: west Bohemia\\
\\
Account ID: 485, District Name: Kutna Hora, District Region: central Bohemia\\
\\
Account ID: 539, District Name: Rakovnik, District Region: central Bohemia\\
\\
Account ID: 1050, District Name: Hodonin, District Region: south Moravia\\
\\
Account ID: 1286, District Name: Zdar nad Sazavou, District Region:\\
\\
south Moravia\\
\\
Account ID:\\
\\
1527,\\
\\
District Name:\\
\\
Tachov,\\
\\
District Region:\\
\\
west Bohemia\\
\\
<...TRUNC>\\
\\
\\
\\
\\
Account ID:\\
1965,\\
District Name:\\
Hodonin,\\
District Region:\\
\\
south Moravia\\
\\
Account ID:\\
\\
2137,\\
\\
District Name:\\
\\
Kladno,\\
\\
District Region:\\
\\
central Bohemia\\
\\
\\
 \\
 Account\\
 \\
ID:\\
\\
\\
2464,\\
\\
\\
\\
\\
<...TRUNC>
\end{quote}

There were multiple new lines between lines above randomly distributed, which were removed from the above response to enhance readability of the text.

This was seen more predominantly in large result-set sizes.

\subsection{Evaluation Results using GPT-4o as the Judge Model}
\label{sec:gpt4oresdets}

Table~\ref{tab:eval0} presents how well different NLR evaluation methods align with human assessments of NLRs. It includes detailed scores for recall, precision, and F1, both at the class level and averaged (macro level). The evaluation methods covered are Metrics-as-a-Judge, LLM-as-a-Judge, and Combo-Eval. This table offers broader results than Table~\ref{tab:eval}, which is limited to F1 scores, by presenting a wider range of statistical metrics.

With GPT-4o as the judge model, Combo-Eval achieves only modest performance gains compared to the LLM-judge approach. Nonetheless, it maintains its overall performance while requiring fewer LLM calls. Conversely, as shown in Table~\ref{tab:evaljudgelms}, when other judge models are used, Combo-Eval offers greater performance gains while simultaneously reducing the number of LLM calls. The reduction in LLM calls remains constant regardless of which judge LLM is being used.

\begin{table*}
	\begin{center}
	\begin{tabular}{|l|c|c|c|c|c|c|}
		\hline
		& \multicolumn{2}{c|}{Metrics-judge} & \multicolumn{2}{|c|}{LLM-judge} & \multicolumn{2}{|c|}{Combo-Eval}  \\
		\hline
		& C0 & C1 & C0 & C1 & C0 & C1  \\
		\hline
            \multicolumn{7}{|l|}{GT} \\
        
		Recall & 46.9 & 89.69 & $65.78\pm$1.4 & $92.92\pm$0.7 & {$71.29\pm$1.0} & {$89.26\pm$0.2} \\
		Prec   & 61.63 & 82.95 & $76.65\pm$1.7 & $88.49\pm$0.4 & {$75.26\pm$0.5} & {$87.85\pm$0.3} \\
		F1     & 53.27 & 86.19 & $70.79\pm$1.1 & $90.65\pm$0.4 & {$73.22\pm$0.7} & {$88.55\pm$0.2} \\
		Rec(macro) &  \multicolumn{2}{c|}{68.30} &  \multicolumn{2}{|c|}{$79.35\pm$0.7} &  \multicolumn{2}{|c|}{$80.27\pm$0.5} \\
		Prec(macro) &  \multicolumn{2}{c|}{72.29} & \multicolumn{2}{|c|}{$82.57\pm$0.9} & \multicolumn{2}{|c|}{$81.56\pm$0.4}\\
		F1(macro) & \multicolumn{2}{c|}{69.73} & \multicolumn{2}{|c|}{$80.72\pm$0.8} & \multicolumn{2}{|c|}{$80.88\pm$0.4}\\
		Accuracy & \multicolumn{2}{c|}{78.52} & \multicolumn{2}{|c|}{$85.83\pm$0.6} & \multicolumn{2}{|c|}{$85.07\pm$0.3}\\
		\hline

        \multicolumn{7}{|l|}{UQDB} \\
		Recall & 50.44 & 84.06 & $49.56\pm$0.9 & $96.15\pm$0.7 &{$55.05\pm$0.9} & {$93.72\pm$0.6} \\
		Prec   & 53.77 & 84.06 & $82.02\pm$2.4 & $84.37\pm$0.2 &{$78.65\pm$2} & {$85.14\pm$0.2} \\
		F1     & 52.05 & 84.06 & $61.77\pm$0.6 & $89.87\pm$0.3(SD) &{$64.75\pm$0.5} & {$89.22\pm$0.3} \\
		Rec(macro) &  \multicolumn{2}{c|}{67.25} &  \multicolumn{2}{|c|}{$72.85\pm$0.3} &  \multicolumn{2}{|c|}{$74.39\pm$0.3} \\
		Prec(macro) &  \multicolumn{2}{c|}{68.92} & \multicolumn{2}{|c|}{$83.19\pm$1.2} & \multicolumn{2}{|c|}{$81.9\pm$1}\\
		F1(macro) & \multicolumn{2}{c|}{68.06} & \multicolumn{2}{|c|}{$75.82\pm$0.4} & \multicolumn{2}{|c|}{$76.98\pm$0.4}\\
		Accuracy & \multicolumn{2}{c|}{75.29} & \multicolumn{2}{|c|}{$83.99\pm$0.4} & \multicolumn{2}{|c|}{$83.99\pm$0.4}\\
		\hline
	\end{tabular}
		\caption{Results (with standard deviation where applicable across 10 runs) from evaluation methods (ROUGE1-recall with threshold to determine classification, LLM as a judge and Combo-Eval method) across class 0 (C0) and class 1 (C1), and average recall, precision, and F1, along with the overall accuracy. Judge LLM is GPT-4o.}
		\label{tab:eval0}
	\end{center}
\end{table*}

\subsection{Misjudgment by Result Size}
\label{sec:misjudge}
\begin{table}
    \begin{center}
        \begin{tabular}{|c|c|c|c|}
            \hline
             & \multicolumn{3}{c|}{Result size}\\
             & 3-9 & 10-49 & 50-499\\
            \hline
            \multicolumn{4}{|l|}{Metrics-Judge} \\
            GT & 30.44 & 30.44 & 39.13 \\
            UQDB & 40.00 & 32.00 & 28.00\\
            \hline
            \multicolumn{4}{|l|}{LLM-judge} \\
            GT & 22.95 & 34.43 & 42.62 \\
            UQDB & 26.76 & 32.39 & 40.85\\
            \hline
            \multicolumn{4}{|l|}{Combo-Eval}\\
            GT & 24.59 & 32.79 & 42.62 \\
            UQDB & 31.34 & 29.85 & 38.81\\
            \hline
        \end{tabular}
        \caption{Percentage of samples that were falsely judged by LLM-judge method broken down by result size.}
        \label{tab:ressizebreak}
    \end{center}
\end{table}

\begin{figure}
    \centering
    \includegraphics[width=1.0\linewidth]{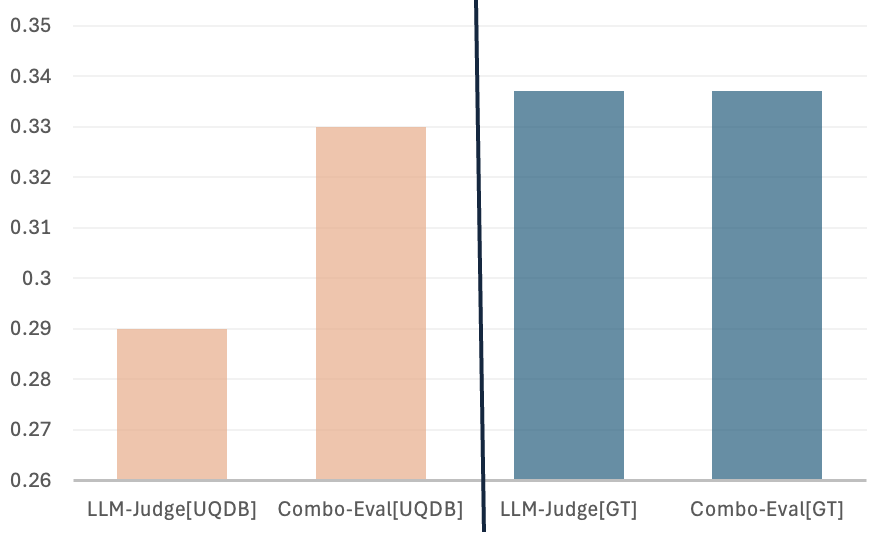}
    \caption{Percentage reduction in incorrect NLR judgments for LLM-judge and Combo-Eval methods compared to Metrics-based evaluation method in UQDB (left) and GT (right) scenarios.}
    \label{fig:incorrect_nlr_reduction}
\end{figure}

Table~\ref{tab:ressizebreak} presents the distribution of falsely judged NLRs categorized by the size of the results. As demonstrated in Figure~\ref{fig:incorrect_nlr_reduction}, both LLM-judge and Combo-Eval methods lead to a reduction in the overall number of incorrect judgments compared to the traditional metrics-based evaluation. However, among the incorrect judgments that remain for each method, Table~\ref{tab:ressizebreak} details the percentage breakdown by result size.

The LLM-as-a-judge method consistently encounters greater difficulty in accurately judging larger result sizes compared to smaller ones.

\subsection{Judgment by Different Judge Models}
\label{sec:judges}

\begin{table*}
	\begin{center}
	\begin{tabular}{|c|c|c|c|c||c|c|c|c|c|c|}
		\hline
		Judge model & \multicolumn{4}{c||}{GT} & \multicolumn{4}{c|}{UQDB} \\
		\hline
		& \multicolumn{2}{|c|}{LLM-as-a-judge} & \multicolumn{2}{|c||}{Combo-Eval \textbf{\textit{(ours)}}} & \multicolumn{2}{|c|}{LLM-as-a-judge} & \multicolumn{2}{|c|}{Combo-Eval \textbf{\textit{(ours)}}} \\
        \hline
		
        L3.3-70b & \multicolumn{2}{|c|}{\color{blue}\textbf{84.60}} & \multicolumn{2}{|c||}{\color{blue}83.90} & \multicolumn{2}{|c|}{77.32} & \multicolumn{2}{|c|}{\textbf{79.40}} \\
        L3.1-405b & \multicolumn{2}{|c|}{\textbf{84.18}} & \multicolumn{2}{|c||}{83.08} & \multicolumn{2}{|c|}{77.79} & \multicolumn{2}{|c|}{\textbf{79.14}} \\
        L3.1-70b & \multicolumn{2}{|c|}{82.93} & \multicolumn{2}{|c||}{\textbf{83.17}} & \multicolumn{2}{|c|}{77.01} & \multicolumn{2}{|c|}{\textbf{79.5}} \\
        O1 & \multicolumn{2}{|c|}{82.68} & \multicolumn{2}{|c||}{\textbf{83.00}} & \multicolumn{2}{|c|}{\color{blue}82.08} & \multicolumn{2}{|c|}{\color{blue}\textbf{82.68}}\\
        Grok-3 & \multicolumn{2}{|c|}{82.43} & \multicolumn{2}{|c||}{\textbf{82.80}} & \multicolumn{2}{|c|}{77.89} & \multicolumn{2}{|c|}{\textbf{78.99}} \\
        Cmd-A & \multicolumn{2}{|c|}{82.31} & \multicolumn{2}{|c||}{\textbf{83.17}} & \multicolumn{2}{|c|}{66.86} & \multicolumn{2}{|c|}{\textbf{70.18}} \\
        GPT-4.1 mini & \multicolumn{2}{|c|}{81.65} & \multicolumn{2}{|c||}{\textbf{83.44}} & \multicolumn{2}{|c|}{77.65} & \multicolumn{2}{|c|}{\textbf{79.90}} \\
        GPT-4o & \multicolumn{2}{|c|}{80.72} &  \multicolumn{2}{|c||}{\textbf{80.88}} & \multicolumn{2}{|c|}{75.82} & \multicolumn{2}{|c|}{\textbf{76.98}} \\
        O3-mini & \multicolumn{2}{|c|}{80.71} & \multicolumn{2}{|c||}{\textbf{81.76}} & \multicolumn{2}{|c|}{75.94} & \multicolumn{2}{|c|}{\textbf{77.79}} \\
        GPT-4.1 & \multicolumn{2}{|c|}{79.75} & \multicolumn{2}{|c||}{\textbf{80.91}} & \multicolumn{2}{|c|}{72.67} & \multicolumn{2}{|c|}{\textbf{74.42}}\\
        CMD-RPlus & \multicolumn{2}{|c|}{77.08} & \multicolumn{2}{|c||}{\textbf{\color{red}78.95}} & \multicolumn{2}{|c|}{65.29} & \multicolumn{2}{|c|}{\textbf{67.48}} \\
        CMD-R & \multicolumn{2}{|c|}{76.02} & \multicolumn{2}{|c||}{\textbf{80.57}} & \multicolumn{2}{|c|}{68.70} & \multicolumn{2}{|c|}{\textbf{72.0}} \\
        GPT-4.1 nano & \multicolumn{2}{|c|}{75.66} & \multicolumn{2}{|c||}{\textbf{82.04}} & \multicolumn{2}{|c|}{\color{red}57.95} & \multicolumn{2}{|c|}{\textbf{\color{red}62.14}} \\
        GPT-4o mini & \multicolumn{2}{|c|}{74.56} & \multicolumn{2}{|c||}{\textbf{\color{red}78.95}} & \multicolumn{2}{|c|}{66.12} & \multicolumn{2}{|c|}{\textbf{69.97}} \\
        Grok-3-mini & \multicolumn{2}{|c|}{\color{red}73.37} & \multicolumn{2}{|c||}{\textbf{79.17}} & \multicolumn{2}{|c|}{70.56} & \multicolumn{2}{|c|}{\textbf{74.61}} \\
		\hline
        Average & \multicolumn{2}{|c|}{79.91} & \multicolumn{2}{|c||}{\textbf{81.72}} & \multicolumn{2}{|c|}{72.64} & \multicolumn{2}{|c|}{\textbf{75.01}} \\
		\hline
	\end{tabular}
		\caption{F1 macro scores across judge LLMs under GT and UQDB scenarios. Best score amongst different judge LLMs is in blue and worst score is in red. Best score between LLM-judge and Comb-eval for each judge model is in bold. Score for Metric-judge is 69.73\% for GT and 68.06\% for UQDB. Results are ordered (descending) by scores on LLM-judge with GT reference.}
		\label{tab:evaljudgelms}
	\end{center}
\end{table*}

Table~\ref{tab:evaljudgelms} shows F1 macro scores across 15 judge LLMs for GT and UQDB scenarios for the LLM-as-a-judge and Combo-Eval methods. The judge LLMs include both closed-source and open-source models of different sizes.

\begin{itemize}
    \item Llama series instruct models (3.3-70B, 3.1-70B, and 3.1-405B)
    
    \item OpenAI GPT-4o~\citep{gpt4o}, 4o mini, 4.1, 4.1 mini, 4.1 nano
    
    \item OpenAI advance reasoning models including O1~\citep{o1} and O3 mini~\citep{o3mini}
    
    \item Cohere Cmd-A~\citep{cmda}, R~\citep{cmdr}, and R+~\citep{cmdrp}
    
    \item Grok-3 and Grok-3-mini~\citep{grok}
\end{itemize}

Combo-Eval outperforms LLM-judge for most of the judge models considered. \textbf{\textit{The results show an average improvement of 1.81\% using Combo-Eval over LLM-judge when GT is used as the reference, and 2.37\% when UQDB is used as the reference.}}

When ground truth NLRs are unavailable, relying on the UQDB approach can be a viable alternative. Otherwise, GT scenario exhibits betters alignment with human assessment of NLRs compared to UQDB. 

The results indicate that, \textbf{\textit{in the UQDB scenario, the alignment with human assessment decreases by 7.27\% for the LLM-judge method compared to using GT. For the Combo-Eval method, the alignment reduces by 6.71\% under the same conditions.}}

\subsection{Evaluation Methods for Determining Accuracy Across Different NLR-Generation LLMs}
\label{sec:genllms}

In the paper, we conducted a human evaluation of NLR generations from different LLMs. We previously shared the agreement between human evaluation of LLM-generated NLRs and the NLR evaluation using three automated evaluation methods - Metric-judge, LLM-judge, and Combo-Eval. Now, we take a different route and compare the overall accuracy evaluation we obtain using Metric-judge, LLM-judge, and Combo-Eval for NLRs produced by the different generation LLMs, and compare that to the overall accuracy we obtain from human evaluation of NLRs generated by each of the different LLMs. This shows how well we can rank different LLMs for the task of generating NLR without human evaluation knowledge. For this exercise, we use GPT-4o as our judge model.

We run the evaluation methods against the test data to determine overall accuracy percentage rather than sample-by-sample alignment with human evaluation. In other words, we calculate accuracy solely based on these evaluation methods, without comparing how well they agree with human assessments for each specific sample. 

The results from this exercise are shared in Table~\ref{tab:llmevalvshuman}. Our findings reveal that all the evaluation methods rank the performance of generation LLMs similarly for the task of NLR generation in a way that matches the rankings from human evaluations. Therefore, to determine which model performs better at generating NLRs, metrics thresholding can be a practical alternative. Although there is a greater discrepancy between human and metrics-based judgments of individual NLRs, this approach can still effectively identify the general trend or ranking of LLMs for this task. 

In an industry setting, this approach can aid in selecting the most suitable model for the task from the available options, thereby supporting critical decision-making and development. 

\begin{table}
	\begin{center}
		\begin{tabular}{|l|c|c|c|c|}
			\hline
			  Judge & Phind & L3.1 70 & L3.1 405 \\
			\hline
			Human & 0.65 & 0.75 & 0.78 \\
                \hline
                \multicolumn{4}{|l|}{GT}\\
                Metrics-Judge & 0.79 & 0.80 & 0.81 \\
                LLM-Judge & 0.72 & 0.8 & 0.82 \\
                Combo-Eval & 0.73 & 0.80 & 0.81 \\
                \hline
                \multicolumn{4}{|l|}{UQDB}\\
                
                Metrics-Judge & 0.73 & 0.76 & 0.76 \\
                LLM-Judge & 0.75 & 0.86 & 0.89 \\
                Combo-Eval & 0.68 & 0.78 & 0.79 \\
			\hline
		\end{tabular}
		\caption{Accuracy of NLRs generated across LLMs using Metrics as a judge, LLM as a judge, Combo-Eval, and human evaluation. Phind=Phind-CodeLlama-34B-v2; L3.170=Llama 3.1-70B-instruct; L3.1405=Llama 3.1-405B-instruct.}
		\label{tab:llmevalvshuman}
	\end{center}
\end{table}

\subsection{Temperature Change for Judge LLMs}
\label{sec:tempjudge}
On a subset of judge LLMs, we ran evaluations using three temperature settings, 0.01, 0.5, and 1. Table~\ref{tab:judgellmtemp} shows the F1 macro score (averaged F1 score representing class 0 and class 1 alignment with human evaluation) for LLM-as-a-judge evaluation method and the Combo-Eval method.

For the Llama-3.3-70B-Instruct model, we observe that lower temperature settings yield slightly better scores across both LLM-as-a-judge and Combo-eval methods under GT as well as UQDB scenarios. On the other hand, for GPT-4o, higher temperature settings result in marginally improved scores, with temperature 0.5 showing highest scores for GT scenario and temperature of 1.0 showing highest scores for UQDB scenario. The other judge models don't exhibit any clear trends. Also, the differences between scores across the various temperature settings across the judge models remain very minimal.

In conclusion, no clear trend emerges from this experiment; the evaluation does not significantly worsen or improve with an increase or decrease in temperature for the task of judging NLRs. The slight variations observed in the results across different temperature settings indicate varying trends for different judge LLMs, suggesting no consistent trend regarding temperature settings across all LLMs. The results imply that different judge models may react differently to temperature changes, but the impact is minimal. Combo-Eval remains superior to LLM-judge across most judge LLMs, and this trend persists across different temperature settings.

\begin{table}
	\begin{center}
		\begin{tabular}{|p{1.3cm}|p{1.9cm}|p{0.7cm}|p{0.7cm}|p{0.7cm}|}
			\hline
			  Judge & Eval method & t0.01 & t0.5 & t1.0\\
			\hline
                \multicolumn{5}{|l|}{GT}\\
                L3.370 & LLM-judge & \textbf{84.60} & 84.25 & 83.39 \\
                L3.370 & Combo-Eval & \textbf{83.90} & 83.53 & 83.17 \\
                Grok-3 & LLM-judge & 82.43 & \textbf{83.45} & 82.58 \\
                Grok-3 & Combo-Eval & \textbf{82.80} & \textbf{82.80} & 82.43 \\
			GPT-4o & LLM-judge & 80.72 & \textbf{81.14} & 80.89 \\
                GPT-4o & Combo-Eval & 80.88 & \textbf{82.00} & 81.88 \\
                G4.1nano & LLM-judge & 75.66 & \textbf{75.88} & 74.37\\
                G4.1nano & Combo-Eval & \textbf{82.04} & 81.69 & 79.61\\
                \hline
                \multicolumn{5}{|l|}{UQDB}\\
                L3.370 & LLM-judge & \textbf{77.32} & 76.24 & 76.18 \\
                L3.370 & Combo-Eval & \textbf{79.40} & 78.30 & 78.28 \\
                Grok-3 & LLM-judge & \textbf{77.89} & 77.08 & 77.71 \\
                Grok-3 & Combo-Eval & 78.99 & 78.85 & \textbf{79.40} \\
			GPT-4o & LLM-judge & 75.82 & 76.54 & \textbf{77.58} \\
                GPT-4o & Combo-Eval & 76.98 & 77.67 & \textbf{78.25} \\
                G4.1nano & LLM-judge & 57.95 & \textbf{58.17} & 54.17\\
                G4.1nano & Combo-Eval & 62.14 & \textbf{62.97} & 59.12\\
			\hline
		\end{tabular}
		\caption{F1 macro score (averaged score from class 1 and class 0 F1) showing alignment of evaluation methods with human judgment across different temperature settings. L3.370=Llama 3.3-70B-instruct; G4.1nano=GPT4.1-nano}
		\label{tab:judgellmtemp}
	\end{center}
\end{table}

\end{document}